\pdfoutput=1

\documentclass[11pt]{article}

\usepackage[]{acl}

\usepackage{times}
\usepackage{latexsym}

\usepackage[T1]{fontenc}

\usepackage[utf8]{inputenc}

\usepackage{microtype}
\usepackage{amsmath}
\usepackage{amsthm}
\usepackage{amssymb}
\usepackage{multirow}
\usepackage{tikz}

\title{An Efficient End-to-End Transformer with Progressive Tri-modal Attention for Multi-modal Emotion Recognition}

\author{
	\bf Yang Wu ~ Pai Peng ~ Zhenyu Zhang ~ Yanyan Zhao\thanks{~~~Corresponding Author} ~~ Bing Qin \\
	Harbin Institute of Technology   \\
	 \{\tt ywu, ppeng, zyzhang, yyzhao, qinb\}@ir.hit.edu.cn \\
}

\begin{document}
\maketitle
\begin{abstract}
Recent works on multi-modal emotion recognition move towards end-to-end models, which can extract the task-specific features supervised by the target task compared with the two-phase pipeline. However, previous methods only model the feature interactions between the textual and either acoustic and visual modalities, ignoring capturing the feature interactions between the acoustic and visual modalities. In this paper, we propose the multi-modal end-to-end transformer (ME2ET), which can effectively model the tri-modal features interaction among the textual, acoustic, and visual modalities at the low-level and high-level. At the low-level, we propose the progressive tri-modal attention, which can model the tri-modal feature interactions by adopting a two-pass strategy and can further leverage such interactions to significantly reduce the computation and memory complexity through reducing the input token length. At the high-level, we introduce the tri-modal feature fusion layer to explicitly aggregate the semantic representations of three modalities. The experimental results on the CMU-MOSEI and IEMOCAP datasets show that ME2ET achieves the state-of-the-art performance. The further in-depth analysis demonstrates the effectiveness, efficiency, and interpretability of the proposed progressive tri-modal attention, which can help our model to achieve better performance while significantly reducing the computation and memory cost. Our code will be publicly available.
\end{abstract}

\section{Introduction}

Multi-modal emotion recognition aims at detecting emotion in the utterance, which consists of three modal inputs, including textual, visual, and acoustic. It has gained increasing attention from people since it is vital for many downstream applications such as natural human-machine interaction and mental healthcare. 

\begin{figure}[tbp]
\centering
\includegraphics[width=0.9\columnwidth]{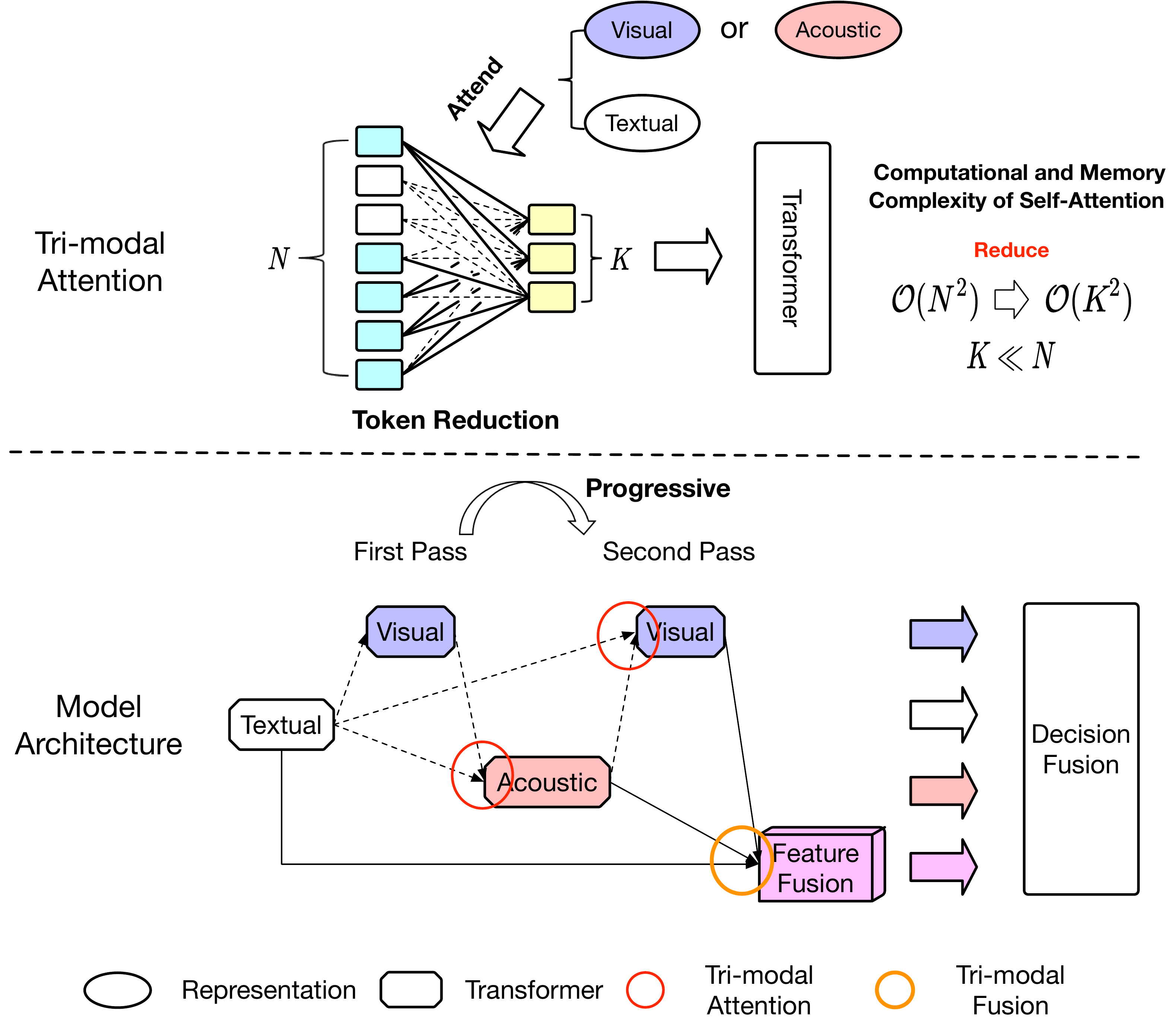} 
\caption{Illustration of ME2ET and the proposed tri-modal attention. Our proposed tri-modal attention can reduce the complexity of self-attention from $\mathcal{O}(N^2)$ to $\mathcal{O}(K^2)$, which makes ME2ET memory and computation-efficient.} 
\label{open}
\end{figure}

Most of the existing studies~\cite{liu-etal-2018-efficient,tsai-etal-2019-multimodal,rahman-etal-2020-integrating} generally use a two-phase pipeline, first extracting uni-modal features from input data and then performing the proposed multi-modal feature fusion method on the features. Different from this line of works, some works train the feature extraction and multi-modal feature fusion modules together in an end-to-end manner since they consider that this approach enables the model to extract the task-specific features supervised by the emotion prediction loss. Dai \emph{et al.}~\shortcite{dai-etal-2021-multimodal} propose the MESM model, which takes the CNN blocks as the feature extraction module and uses the cross-modal attention to capture bi-modal feature interactions. However, MESM only captures the features interactions between the textual and either acoustic or visual modalities, ignoring leveraging the feature interactions between the acoustic and visual modalities, which is very useful for understanding the sentiment semantics~\cite{ zadeh-etal-2017-tensor, tsai-etal-2019-multimodal}.

To address this problem, we propose the multi-modal end-to-end transformer (ME2ET) shown in Figure~\ref{open}, which can capture the tri-modal feature interactions among the textual, acoustic, and visual modalities effectively at the low-level and high-level. At the high-level, we propose the tri-modal feature fusion layer to sufficiently fuse the three uni-modal semantic representations. At the low-level, we propose the progressive tri-modal attention. The main idea of it is \textit{to generate fewer but more informative visual/acoustic tokens based on the input visual/acoustic tokens by leveraging the tri-modal feature interactions.} Through reducing the length of the input tokens, we can significantly reduce the computation and memory complexity. If the length of the input tokens is $N$ and the length of the generated tokens is $K$($K{\ll}N$), which is a hyper-parameter, we can reduce the computation and memory complexity of the self-attention block in the visual/acoustic transformer from $\mathcal{O}(N^2)$ to $\mathcal{O}(K^2)$. More details about the complexity  analysis are given in Appendix \ref{complexity}.

To be more specific, we adopt a simple two-pass strategy to capture the tri-modal feature interactions in the proposed progressively tri-modal attention. In the first pass, considering the visual data contains more noise as some parts of the face may be missing, we utilize the textual representation to attend the input visual tokens and generate fewer tokens, which are then passed into the visual transformer and get the preliminary visual representation. In the second pass, we use not only the textual but also visual representations to attend the acoustic tokens and feed the outputs into the acoustic transformer producing the acoustic representation. Subsequently, we perform the attention mechanism on the original visual tokens again using both the textual and acoustic representations and get the final visual representation. In this way, our model can obtain the semantic uni-modal representations effectively by leveraging the tri-modal feature interactions.

We conduct extensive experiments on CMU-MOSEI~\cite{bagher-zadeh-etal-2018-multimodal} and IEMOCAP~\cite{busso2008iemocap}. The experimental results show that our model surpasses the baselines and achieves the state-of-the-art performance, which demonstrates the effectiveness of our model. We further analyze the contribution of each part of our model, and the results indicate that the progressive tri-modal attention and tri-modal feature fusion layer are necessary for our model, which can capture the tri-modal feature interactions at the low-level and high-level. Moreover, the in-depth analysis of the progressive tri-modal attention including the ablation study, computation and memory analysis, and visualization analysis shows its effectiveness, efficiency, and interpretability.

The main contributions of this work are as follows:
\begin{itemize}
\item We propose the progressive tri-modal attention, which can help our model to achieve better performance while significantly reducing the computation and memory cost by fully exploring the tri-modal feature interactions.
\item We introduce the multi-modal end-to-end transformer (ME2ET), which is a simple and efficient \emph{purely} transformer-based multi-modal emotion recognition framework. To facilitate further research, we explore different patch construction methods for the acoustic and visual raw data and analyze their effects. 
\item We evaluate ME2ET on two public datasets and ME2ET obtains the state-of-the-art results. We also conduct an in-depth analysis to show the effectiveness, efficiency and interpretability of ME2ET.
\end{itemize} 

\section{Related Work}
\subsection{Multi-modal Emotion Recognition}
There are two lines of works conducted on utterance-level multi-modal emotion recognition. One line of works adopts a two-phase pipeline, first extracting features and then fusing them. MulT~\cite{tsai-etal-2019-multimodal} uses the cross-modal attention to capture the bi-modal interactions. EmoEmbs~\cite{dai-etal-2020-modality} leverages the cross-modal emotion embeddings for multi-modal emotion recognition. ~\citet{lv2021progressive} introduce the message hub to explore the inter-modal interactions. The other line of works trains the whole model in an end-to-end manner since they consider the extracted features may not be suitable for the target task and can not be fine-tuned, which may lead to sub-optimal performance. MESM~\cite{dai-etal-2021-multimodal} applies the VGG blocks to extract the visual and acoustic features and proposes the cross-modal attention to make the model focus on the important features. In this paper, we focus on the end-to-end multi-modal emotion recognition and propose the multi-modal end-to-end transformer (ME2ET). There are mainly three differences between our model and MESM. (1) MESM fails to leverage the feature interactions between the visual and acoustic modalities. To fully leverage tri-modal information, we propose the progressive tri-modal attention and tri-modal feature fusion layer to capture the tri-modal feature interactions; (2) we directly apply the vision transformer to process the raw audio and video data while MESM utilizes CNNs as the encoders considering its strong capacity of capturing the global intra-modal dependencies. Moreover, our model can easily benefit from the emerging vision transformer field, such as adopting stronger pre-trained transformers to boost model performance; (3) the proposed progressive tri-modal attention is different from the bi-modal attention adopted by MESM. The main idea of our proposed attention is to generate fewer but more useful tokens based on the input tokens. But the core idea of the cross-modal attention utilized by MESM is masking the less important area.

\subsection{Transformer Network}
Inspired by the recent successes of Transformer in NLP, multiple works try utilizing Transformer for image understanding.  ViT~\cite{dosovitskiy2021an} is the first work attempting to apply a standard Transformer directly to images, which first splits an image into patches and then passes them to the transformer model. DeiT~\cite{touvron2021training} takes a convnet model as the teacher and learns knowledge from it by reproducing the label predicted by the teacher. Besides, the vision transformer can also be adapted for audio classification~\cite{gong21b_interspeech}. Considering the huge computation and memory cost of Transformer, some transformer variants are proposed. Reformer~\cite{Kitaev2020Reformer} replaces dot-product attention in self-attention by one that uses locality-sensitive
hashing and reduces the time and memory complexity. ~\citet{ryoo2021tokenlearner} insert the TokenLearner layer into the middle of the vision transformer to reduce the token number. T2T-ViT~\cite{yuan2021tokens} utilizes a layer-wise Tokens-to-Token module to model the local structure information of the input image, which also reduces the length of tokens. TR-BERT~\cite{ye-etal-2021-tr} learns to dynamically select important tokens.

Our work, in contrast to these works, \textit{explicitly leverages the tri-modal feature interaction} to transform a long sequence of tokens into fewer but more informative tokens, which not only enhances the model performance but also significantly reduces the computation and memory cost.

\begin{figure*}[t]
\centering
\includegraphics[width=1.8\columnwidth]{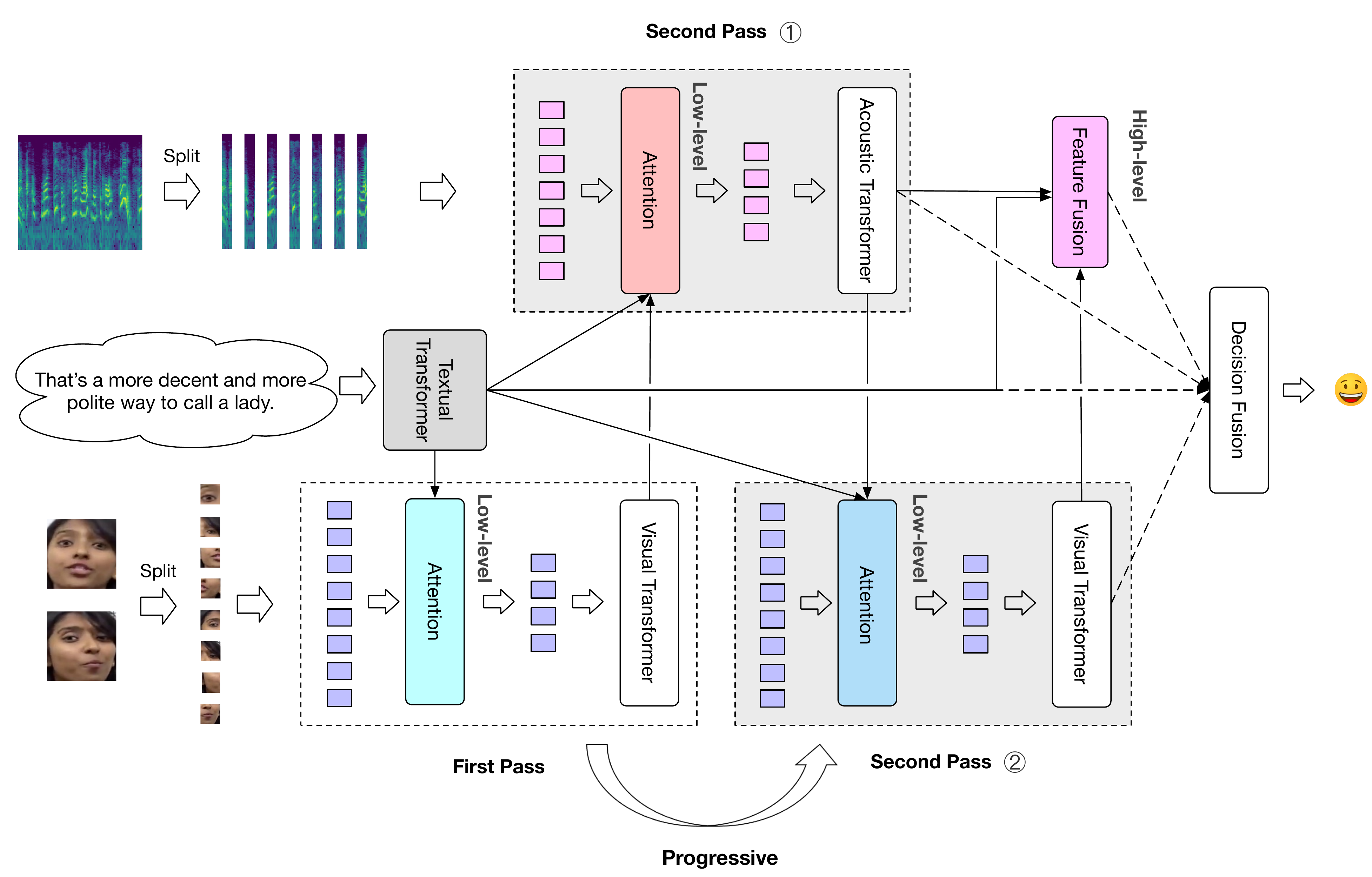} 
\caption{An illustration of the proposed ME2ET model. ME2ET uses the progressive tri-modal attention and the tri-modal feature fusion layer to fully explore the tri-modal feature interactions at the low-level and high-level.}
\label{model}
\end{figure*}

\section{Approach}
In this section, we introduce the proposed multi-modal end-to-end transformer (ME2ET) in detail, which is shown in Figure \ref{model}. ME2ET adopts the progressive tri-modal attention to leverage the tri-modal feature interactions at the low-level to generate fewer but more useful tokens based on the input tokens, which significantly reduces the memory and computation cost. In addition, ME2ET also uses the tri-modal feature fusion layer to model the tri-modal feature interactions at the high-level.

\subsection{Background} \label{background}
We first introduce the transformer architecture~\cite{vaswani2017attention}, which consists of two basic components: multi-head self-attention (MSA) and feed-forward network (FFN). 

\paragraph{MSA} Given an input token sequence $\mathbf{x} \in \mathbb{R}^{l \times d}$, we first obtain the queries, keys, and values by Equation \ref{eqn:1}. And then, for each element, we compute the similarity between it and all tokens and obtain the attention weights. Finally, we apply the weights to the values and get the outputs.
\begin{equation}
\label{eqn:1}
[\mathbf{q}, \mathbf{k} ,\mathbf{v}] = \mathbf{x}W_{qkv}
\end{equation}
\begin{equation}
\label{eqn:2}
\text{SA}(\mathbf{x}) = \text{softmax}(\frac{\mathbf{q}\mathbf{k}^\top}{\sqrt{d_h}})\mathbf{v}
\end{equation}
where $W_{qkv} \in \mathbb{R}^{d \times 3d_h}$.

MSA is an extension of the self-attention (SA) introduced above. Specifically, We run $k$ different self-attention functions, and project their concatenated outputs, resulting in the final results.
\begin{equation}
\label{eqn:3}
\text{MSA}(\mathbf{x}) = [\text{SA}_1(\mathbf{x});\cdots;\text{SA}_k(\mathbf{x})] W_{msa}
\end{equation}
where $W_{msa} \in \mathbb{R}^{k\cdot{d_h} \times d}$.

\paragraph{FFN} FFN consists of two linear layers with a GELU ~\cite{hendrycks2016gaussian} activation in between. Given an input sequence $\mathbf{z} \in \mathbb{R}^{l \times d}$, the outputs are calculated by Equation \ref{eqn:4}.
\begin{equation}
\label{eqn:4}
\text{FFN}(\mathbf{z}) = \text{GELU}(\mathbf{z}W_{1} + b_1)W_{2} + b_2
\end{equation}
where $W_{1} \in \mathbb{R}^{d \times 4\cdot{d}}$, $W_{2} \in \mathbb{R}^{ 4\cdot{d} \times d}$, $b_{1} \in \mathbb{R}^{4\cdot{d}}$, and $b_{2} \in \mathbb{R}^{d}$.

The transformer architecture can be described as follows\footnote{To obtain the whole representation of the input, we append [CLS] to the beginning following BERT and ViT. As for the textual transformer, the segmentation embeddings are also added to the input token embeddings.}.
\begin{equation}
\begin{aligned}
\mathbf{z}_0 &= [\mathbf{x}_{cls};\mathbf{t}] + \mathbf{e}_{pos} \\
{\mathbf{z}}^{'}_{l} &= \text{MSA}(\text{LN}(\mathbf{z}_{l-1})) + \mathbf{z}_{l-1} \\
\mathbf{z}_{l} &= \text{FFN}(\text{LN}({\mathbf{z}}^{'}_{l})) + {\mathbf{z}}^{'}_{l} \\
\end{aligned}
\end{equation}
where $\mathbf{t}$, $\mathbf{e}_{pos}$, and LN denote the input tokens, the position embeddings, and the LayerNorm layer~\cite{ba2016layer} respectively.

We can use the transformer model to encode the input token embeddings $\mathbf{t}$, producing $\mathbf{z}_{L}$, where $L$ is the layer number. Generally, we take the representation of $\mathbf{x}_{cls}$ as the output.
\begin{equation}
\mathbf{z}_{L} = \text{Transformer}(\mathbf{t}) \\
\end{equation}

\subsection{Token Construction} \label{token_section}

\paragraph{Visual Tokens} To apply the transformer architecture to the visual input, we transform the raw data into tokens, which can be passed to the transformer model. Given a sequence of face images $\mathbf{x^v} \in \mathbb{R}^{J \times H \times W \times C}$, we split each image in them into a sequence of patches $\mathbf{x^v_p} \in \mathbb{R}^{P \times P \times C}$ without overlap, and then concatenate the patches, resulting in $Q$ patches, where $Q = JHW/P^2$ is the number of patches.  $J$, $H$, $W$, and $C$ are the image number, the height of the face image, the width of the face image, and the channel number of the face image respectively. We then map each patch into a 1D token embedding using a linear projection layer, producing $Q$ tokens $\mathbf{t^v} \in \mathbb{R}^{Q \times d}$.  

\paragraph{Acoustic Tokens} For the acoustic input, we first convert the input audio waveform into the spectrogram $\mathbf{x^a} \in \mathbb{R}^{F \times T \times 1}$ and then split it into rectangular patches $\mathbf{x^a_p} \in \mathbb{R}^{F \times 2 \times 1}$  in the temporal order to keep the timing information, where $F$ is the dimension of features and $T$ is the frame number. Finally, we apply a linear layer to transform the patches into tokens, producing $M$ tokens $\mathbf{t^a} \in \mathbb{R}^{M \times d}$ .

\paragraph{Textual Tokens} We construct the textual tokens following BERT~\cite{kenton2019bert}. We then pass the textual tokens to BERT and obtain the representation of [CLS], denoted as $v_l$.

\subsection{Progressive Tri-modal Attention} \label{trimodal_section}
To capture tri-modal feature interactions at the low-level, we propose the progressive tri-modal attention, which consists of two passes. In the first pass, we obtain the preliminary visual representation by leveraging the textual information. Note that, we do not directly take this representation as the final visual representation, as we consider that incorporating the acoustic information can further improve the visual representation. Specifically, the preliminary visual representation is obtained as follows. Given the visual tokens $\mathbf{t^v} \in \mathbb{R}^{Q \times d}$ and acoustic tokens $\mathbf{t^a} \in \mathbb{R}^{M \times d}$, instead of passing the transformed tokens directly into the transformer block, we first use the proposed attention mechanism to generate new visual tokens $\mathbf{z^{v}} \in \mathbb{R}^{K \times d}$ leveraging the textual information and the token number $K$ is a hyper-parameter and is smaller than $Q$ and $M$. This approach can significantly reduce the computation and memory complexity. Then we pass the $\mathbf{z^{v}}$ to the visual transformer and obtain the preliminary visual representation $v_{one}$. 
\begin{equation}
\label{eqn:5}
\begin{aligned}
\mathbf{z^{v}} &= \text{softmax}([\mathbf{t^v};\mathbf{v^l}]W_{lv} + b_{lv})^\top\mathbf{t^v}  \\
v_{one} &= \text{Transformer}_v(\mathbf{z^{v}})
\end{aligned}
\end{equation}
where $\mathbf{v^l}$ is obtained by repeating $v_{l}$, $W_{lv} \in \mathbb{R}^{2 \cdot d \times K}$ and $b_{lv} \in \mathbb{R}^{K}$. 

In the second pass, we utilize the textual representation $v_{l}$ and the preliminary visual representation $v_{one}$ to guide the model to address the important acoustic tokens and obtain the acoustic representation $a$.  
\begin{equation}
\label{eqn:6}
\begin{aligned}
\mathbf{z^{a}} &= \text{softmax}([\mathbf{t^a};\mathbf{v^v};\mathbf{v^l}]W_{la} + b_{la})^\top\mathbf{t^a}  \\
a &= \text{Transformer}_a(\mathbf{z^{a}})
\end{aligned}
\end{equation}
where $\mathbf{v^l}$, $\mathbf{v^v}$ are obtained by repeating $v_{l}$ and $v_{one}$ respectively, $W_{la} \in \mathbb{R}^{3 \cdot d \times K}$, and $b_{la} \in \mathbb{R}^{K}$.

Subsequently, we perform the attention over the original visual tokens again since in the first pass the acoustic information is not utilized, which is useful for selecting informative visual tokens. The generated tokens are then passed into \textbf{the same visual transformer} used in the first pass, producing the refined visual representation $v$. Finally, we get the visual representation $v$ and acoustic representation $a$ by leveraging the tri-modal feature interactions.
\begin{equation}
\label{eqn:7}
\begin{aligned}
\mathbf{z^{\hat{v}}} &= \text{softmax}([\mathbf{t^v};\mathbf{v^a};\mathbf{v^l}]W_{l\hat{v}} + b_{l\hat{v}})^\top\mathbf{t^v}  \\
v &= \text{Transformer}_v(\mathbf{z^{\hat{v}}})
\end{aligned}
\end{equation}
where $\mathbf{v^l}$, $\mathbf{v^a}$ are obtained by repeating $v_{l}$ and $a$ respectively, $W_{l\hat{v}} \in \mathbb{R}^{3 \cdot d \times K}$, and $b_{l\hat{v}} \in \mathbb{R}^{K}$.

\subsection{Multi-modal Feature Fusion} \label{fuse_section}
To capture the tri-modal feature interactions at the high-level, we propose the tri-modal feature fusion layer to fuse the semantic representations obtained by the transformer models and predict the results. Finally, we apply the decision fusion layer to aggregate the predicted results and generate the final prediction label.
\begin{equation}
\label{eqn:8}
\begin{aligned}
p_{fusion} &= [v; a; v_{l}] W_{fusion} + b_{fusion} \\
p_{v} &= vW_{fv} + b_{fv} \\
p_{a} &= aW_{fa} + b_{fa} \\
p_{l} &= v_{l}W_{fl} + b_{fl} \\
p &= ([p_{v}^\top;p_{a}^\top;p_{l}^\top;p_{fusion}^\top]W_{decision})^\top
 \end{aligned}
\end{equation}
where $W_{fv}, W_{fa}, W_{fl} \in \mathbb{R}^{d \times C}$, $W_{fusion} \in \mathbb{R}^{3 \cdot d \times C}$, $W_{decision} \in \mathbb{R}^{4 \times 1}$, $b_{fv}, b_{fa}, b_{fl} \in \mathbb{R}^{C}$, $b_{fusion} \in \mathbb{R}^{C}$, and $C$ is the class number.

\begin{table*}[tbp]
    \caption{Results on the IEMOCAP dataset. \dag indicates the results are copied from~\cite{dai-etal-2021-multimodal}. The best results are in \textbf{bold}.}
	\centering
	\resizebox{0.95\linewidth}{!}{
	\begin{tabular}{l|l|llllllllllllll}
		\hline
		\multicolumn{1}{c|}{\multirow{2}{*}{Strategy}}  &\multicolumn{1}{c|}{\multirow{2}{*}{Models}}  & \multicolumn{2}{c}{Angry}  & \multicolumn{2}{c}{Excited} & \multicolumn{2}{c}{Frustrated} & \multicolumn{2}{c}{Happy} & \multicolumn{2}{c}{Neutral}   
		& \multicolumn{2}{c}{Sad} & \multicolumn{2}{c}{\textbf{Average}}\\
		 & & \multicolumn{1}{c}{Acc.} & \multicolumn{1}{c}{F1} & \multicolumn{1}{c}{Acc.} & \multicolumn{1}{c}{F1}  & \multicolumn{1}{c}{Acc.} & \multicolumn{1}{c}{F1} 
		 & \multicolumn{1}{c}{Acc.} & \multicolumn{1}{c}{F1}  & \multicolumn{1}{c}{Acc.} & \multicolumn{1}{c}{F1} & \multicolumn{1}{c}{Acc.} & \multicolumn{1}{c}{F1} 
		 & \multicolumn{1}{c}{Acc.} & \multicolumn{1}{c}{F1}  \\
		\hline
		\hline
		\multicolumn{1}{c|}{\multirow{4}{*}{Pipeline}}  &	\multicolumn{1}{c|}{LF-LSTM\dag}	 & \multicolumn{1}{c}{71.2} & \multicolumn{1}{c}{49.4} & \multicolumn{1}{c}{79.3} & \multicolumn{1}{c}{57.2}  & \multicolumn{1}{c}{68.2} & \multicolumn{1}{c}{51.5} 
		 & \multicolumn{1}{c}{67.2} & \multicolumn{1}{c}{37.6}  & \multicolumn{1}{c}{66.5} & \multicolumn{1}{c}{47.0} & \multicolumn{1}{c}{78.2} & \multicolumn{1}{c}{54.0} 
		 & \multicolumn{1}{c}{71.8} & \multicolumn{1}{c}{49.5}  \\
		&	\multicolumn{1}{c|}{LF-TRANS\dag}	 & \multicolumn{1}{c}{81.9} & \multicolumn{1}{c}{50.7} & \multicolumn{1}{c}{85.3} & \multicolumn{1}{c}{57.3}  & \multicolumn{1}{c}{60.5} & \multicolumn{1}{c}{49.3} 
		 & \multicolumn{1}{c}{85.2} & \multicolumn{1}{c}{37.6}  & \multicolumn{1}{c}{72.4} & \multicolumn{1}{c}{49.7} & \multicolumn{1}{c}{87.4} & \multicolumn{1}{c}{ 57.4} 
		 & \multicolumn{1}{c}{78.8} & \multicolumn{1}{c}{50.3}  \\
		 & \multicolumn{1}{c|}{EmoEmbs\dag}	 & \multicolumn{1}{c}{65.9} & \multicolumn{1}{c}{48.9} & \multicolumn{1}{c}{73.5} & \multicolumn{1}{c}{58.3}  & \multicolumn{1}{c}{ 68.5} & \multicolumn{1}{c}{52.0} 
		 & \multicolumn{1}{c}{69.6} & \multicolumn{1}{c}{38.3}  & \multicolumn{1}{c}{73.6} & \multicolumn{1}{c}{48.7} & \multicolumn{1}{c}{80.8} & \multicolumn{1}{c}{ 53.0} 
		 & \multicolumn{1}{c}{72.0} & \multicolumn{1}{c}{49.8}  \\
		 
		  & \multicolumn{1}{c|}{MulT\dag}	 & \multicolumn{1}{c}{77.9} & \multicolumn{1}{c}{60.7} & \multicolumn{1}{c}{76.9} & \multicolumn{1}{c}{58.0}  & \multicolumn{1}{c}{72.4} & \multicolumn{1}{c}{57.0} 
		 & \multicolumn{1}{c}{80.0} & \multicolumn{1}{c}{46.8}  & \multicolumn{1}{c}{74.9} & \multicolumn{1}{c}{53.7} & \multicolumn{1}{c}{83.5} & \multicolumn{1}{c}{65.4} 
		 & \multicolumn{1}{c}{77.6} & \multicolumn{1}{c}{56.9}  \\
		 \hline
		 \multicolumn{1}{c|}{\multirow{4}{*}{End-to-End}} & \multicolumn{1}{c|}{FE2E\dag}	 & \multicolumn{1}{c}{88.7} & \multicolumn{1}{c}{63.9} & \multicolumn{1}{c}{89.1} & \multicolumn{1}{c}{61.9}  & \multicolumn{1}{c}{71.2} & \multicolumn{1}{c}{57.8} 
		 & \multicolumn{1}{c}{90.0} & \multicolumn{1}{c}{44.8}  & \multicolumn{1}{c}{79.1} & \multicolumn{1}{c}{58.4} & \multicolumn{1}{c}{89.1} & \multicolumn{1}{c}{65.7} 
		 & \multicolumn{1}{c}{84.5} & \multicolumn{1}{c}{58.8}  \\
		 & \multicolumn{1}{c|}{MESM\dag}	 & \multicolumn{1}{c}{88.2} & \multicolumn{1}{c}{62.8} & \multicolumn{1}{c}{88.3} & \multicolumn{1}{c}{61.2}  & \multicolumn{1}{c}{74.9} & \multicolumn{1}{c}{58.4} 
		 & \multicolumn{1}{c}{89.5} & \multicolumn{1}{c}{\textbf{47.3}}  & \multicolumn{1}{c}{77.0} & \multicolumn{1}{c}{52.0} & \multicolumn{1}{c}{88.6} & \multicolumn{1}{c}{62.2} 
		 & \multicolumn{1}{c}{84.4} & \multicolumn{1}{c}{57.4}  \\
		 
		 & \multicolumn{1}{c|}{FE2E-BERT}	 & \multicolumn{1}{c}{\textbf{90.6}} & \multicolumn{1}{c}{64.4} & \multicolumn{1}{c}{87.3} & \multicolumn{1}{c}{59.9}  & \multicolumn{1}{c}{72.4} & \multicolumn{1}{c}{59.8} 
		 & \multicolumn{1}{c}{90.6} & \multicolumn{1}{c}{43.4}  & \multicolumn{1}{c}{77.3} & \multicolumn{1}{c}{54.0} & \multicolumn{1}{c}{90.0} & \multicolumn{1}{c}{65.2} 
		 & \multicolumn{1}{c}{84.7} & \multicolumn{1}{c}{57.8}  \\
		
		& \multicolumn{1}{c|}{MESM-BERT}	 & \multicolumn{1}{c}{86.6} & \multicolumn{1}{c}{63.0} & \multicolumn{1}{c}{89.2} & \multicolumn{1}{c}{57.6}  & \multicolumn{1}{c}{75.6} & \multicolumn{1}{c}{58.5} 
		 & \multicolumn{1}{c}{\textbf{90.7}} & \multicolumn{1}{c}{43.1}  & \multicolumn{1}{c}{77.5} & \multicolumn{1}{c}{54.6} & \multicolumn{1}{c}{89.2} & \multicolumn{1}{c}{65.3} 
		 & \multicolumn{1}{c}{84.8} & \multicolumn{1}{c}{57.0}  \\

		 \hline

		\multicolumn{1}{c|}{\multirow{1}{*}{End-to-End}}  & \multicolumn{1}{c|}{Ours}	 & \multicolumn{1}{c}{89.8} & \multicolumn{1}{c}{\textbf{65.9}} & \multicolumn{1}{c}{\textbf{89.2}} & \multicolumn{1}{c}{\textbf{63.9}}  & \multicolumn{1}{c}{\textbf{79.2}} & \multicolumn{1}{c}{\textbf{60.7}} & \multicolumn{1}{c}{90.0} & \multicolumn{1}{c}{44.7}  & \multicolumn{1}{c}{\textbf{78.8}} & \multicolumn{1}{c}{\textbf{58.7}} & \multicolumn{1}{c}{\textbf{92.4}} & \multicolumn{1}{c}{\textbf{73.8}} & \multicolumn{1}{c}{\textbf{86.5}} & \multicolumn{1}{c}{\textbf{61.3}}  \\	 
		 \hline
        \end{tabular}
    }
	\label{tab:main_iemocap}
\end{table*}

\begin{table*}[tbp]
    \caption{Results on the CMU-MOSEI dataset. \dag indicates the results are copied from~\cite{dai-etal-2021-multimodal}. The best results are in \textbf{bold}.}
	\centering
	\resizebox{\linewidth}{!}{
	\begin{tabular}{l|l|llllllllllllll}
		\hline
		\multicolumn{1}{c|}{\multirow{2}{*}{Strategy}}  &\multicolumn{1}{c|}{\multirow{2}{*}{Models}}  & \multicolumn{2}{c}{Angry}  & \multicolumn{2}{c}{Disgusted} & \multicolumn{2}{c}{Fear} & \multicolumn{2}{c}{Happy} & \multicolumn{2}{c}{Sad}   
		& \multicolumn{2}{c}{Surprised} & \multicolumn{2}{c}{\textbf{Average}}\\
		 & & \multicolumn{1}{c}{WAcc.} & \multicolumn{1}{c}{F1} & \multicolumn{1}{c}{WAcc.} & \multicolumn{1}{c}{F1}  & \multicolumn{1}{c}{WAcc.} & \multicolumn{1}{c}{F1} 
		 & \multicolumn{1}{c}{WAcc.} & \multicolumn{1}{c}{F1}  & \multicolumn{1}{c}{WAcc.} & \multicolumn{1}{c}{F1} & \multicolumn{1}{c}{WAcc.} & \multicolumn{1}{c}{F1} 
		 & \multicolumn{1}{c}{WAcc.} & \multicolumn{1}{c}{F1}  \\
		\hline
		\hline
		\multicolumn{1}{c|}{\multirow{4}{*}{Pipeline}}  &\multicolumn{1}{c|}{LF-LSTM\dag} & \multicolumn{1}{c}{64.5} & \multicolumn{1}{c}{47.1} & \multicolumn{1}{c}{70.5} & \multicolumn{1}{c}{49.8} & \multicolumn{1}{c}{61.7} & \multicolumn{1}{c}{22.2} & \multicolumn{1}{c}{61.3} & \multicolumn{1}{c}{73.2} & \multicolumn{1}{c}{63.4} & \multicolumn{1}{c}{47.2} & \multicolumn{1}{c}{57.1} & \multicolumn{1}{c}{20.6} & \multicolumn{1}{c}{63.1} & \multicolumn{1}{c}{43.3} \\
		&\multicolumn{1}{c|}{LF-TRANS\dag} & \multicolumn{1}{c}{65.3} & \multicolumn{1}{c}{47.7} & \multicolumn{1}{c}{74.4} & \multicolumn{1}{c}{51.9} & \multicolumn{1}{c}{62.1} & \multicolumn{1}{c}{24.0} & \multicolumn{1}{c}{60.6} & \multicolumn{1}{c}{72.9} & \multicolumn{1}{c}{60.1} & \multicolumn{1}{c}{45.5} & \multicolumn{1}{c}{62.1} & \multicolumn{1}{c}{24.2} & \multicolumn{1}{c}{64.1} & \multicolumn{1}{c}{44.4}  \\
		 &\multicolumn{1}{c|}{EmoEmbs\dag} & \multicolumn{1}{c}{66.8} & \multicolumn{1}{c}{49.4} & \multicolumn{1}{c}{69.6} & \multicolumn{1}{c}{48.7} & \multicolumn{1}{c}{63.8} & \multicolumn{1}{c}{23.4} & \multicolumn{1}{c}{61.2} & \multicolumn{1}{c}{71.9} & \multicolumn{1}{c}{60.5} & \multicolumn{1}{c}{47.5} & \multicolumn{1}{c}{63.3} & \multicolumn{1}{c}{24.0} & \multicolumn{1}{c}{64.2} & \multicolumn{1}{c}{44.2}	 \\
		&\multicolumn{1}{c|}{MulT\dag}	& \multicolumn{1}{c}{64.9} & \multicolumn{1}{c}{47.5} & \multicolumn{1}{c}{71.6} & \multicolumn{1}{c}{49.3} & \multicolumn{1}{c}{62.9} & \multicolumn{1}{c}{25.3} & \multicolumn{1}{c}{\textbf{67.2}} & \multicolumn{1}{c}{\textbf{75.4}} & \multicolumn{1}{c}{64.0} & \multicolumn{1}{c}{48.3} & \multicolumn{1}{c}{61.4} & \multicolumn{1}{c}{25.6} & \multicolumn{1}{c}{65.4} & \multicolumn{1}{c}{45.2}  \\	
		\hline
		 \multicolumn{1}{c|}{\multirow{4}{*}{End-to-End}}  &\multicolumn{1}{c|}{FE2E\dag}	 & \multicolumn{1}{c}{67.0} & \multicolumn{1}{c}{49.6} & \multicolumn{1}{c}{\textbf{77.7}} & \multicolumn{1}{c}{\textbf{57.1}} & \multicolumn{1}{c}{63.8} & \multicolumn{1}{c}{26.8} & \multicolumn{1}{c}{65.4} & \multicolumn{1}{c}{72.6} & \multicolumn{1}{c}{65.2} & \multicolumn{1}{c}{49.0} & \multicolumn{1}{c}{\textbf{66.7}} & \multicolumn{1}{c}{\textbf{29.1}} & \multicolumn{1}{c}{67.6} & \multicolumn{1}{c}{47.4}  \\
		 &\multicolumn{1}{c|}{MESM\dag}	 & \multicolumn{1}{c}{66.8} & \multicolumn{1}{c}{49.3} & \multicolumn{1}{c}{75.6} & \multicolumn{1}{c}{56.4} & \multicolumn{1}{c}{65.8} & \multicolumn{1}{c}{28.9} & \multicolumn{1}{c}{64.1} & \multicolumn{1}{c}{72.3} & \multicolumn{1}{c}{63.0} & \multicolumn{1}{c}{46.6} & \multicolumn{1}{c}{65.7} & \multicolumn{1}{c}{27.2} & \multicolumn{1}{c}{66.8} & \multicolumn{1}{c}{46.8}  \\
		 &\multicolumn{1}{c|}{FE2E-BERT}	 & \multicolumn{1}{c}{66.8} & \multicolumn{1}{c}{49.4} & \multicolumn{1}{c}{72.8} & \multicolumn{1}{c}{55.0} & \multicolumn{1}{c}{66.2} & \multicolumn{1}{c}{29.1} & \multicolumn{1}{c}{66.5} & \multicolumn{1}{c}{71.4} & \multicolumn{1}{c}{64.4} & \multicolumn{1}{c}{48.4} & \multicolumn{1}{c}{62.5} & \multicolumn{1}{c}{28.1} & \multicolumn{1}{c}{66.5} & \multicolumn{1}{c}{46.9} \\
		&\multicolumn{1}{c|}{MESM-BERT}	  & \multicolumn{1}{c}{66.5} & \multicolumn{1}{c}{49.2} & \multicolumn{1}{c}{77.6} & \multicolumn{1}{c}{54.0} & \multicolumn{1}{c}{69.2} & \multicolumn{1}{c}{28.4} & \multicolumn{1}{c}{62.7} & \multicolumn{1}{c}{72.2} & \multicolumn{1}{c}{63.6} & \multicolumn{1}{c}{47.7} & \multicolumn{1}{c}{60.5} & \multicolumn{1}{c}{26.3} & \multicolumn{1}{c}{66.7} & \multicolumn{1}{c}{46.3}  \\

		 \hline
		 \multicolumn{1}{c|}{\multirow{1}{*}{End-to-End}}  & \multicolumn{1}{c|}{Ours}	 & \multicolumn{1}{c}{\textbf{67.9}} & \multicolumn{1}{c}{\textbf{51.1}} & \multicolumn{1}{c}{76.4} & \multicolumn{1}{c}{56.4}  & \multicolumn{1}{c}{\textbf{69.3}} & \multicolumn{1}{c}{\textbf{29.3}} 
		 & \multicolumn{1}{c}{66.4} & \multicolumn{1}{c}{73.2}  & \multicolumn{1}{c}{\textbf{66.2}} & \multicolumn{1}{c}{\textbf{50.0}} & \multicolumn{1}{c}{63.3} & \multicolumn{1}{c}{27.7} 
		 & \multicolumn{1}{c}{\textbf{68.3}} & \multicolumn{1}{c}{\textbf{48.0}}  \\
		 
		 \hline
        \end{tabular}
    }

	\label{tab:main_mosei}
\end{table*}

\section{Experiments}

\subsection{Datasets}
We evaluate our model on CMU-MOSEI~\cite{bagher-zadeh-etal-2018-multimodal} and IEMOCAP~\cite{busso2008iemocap}. We use the reorganized datasets released by the previous work~\cite{dai-etal-2021-multimodal} since the original datasets do not provide the aligned raw data. IEMOCAP consists of 7,380 annotated utterances. Each utterance in this dataset consists of three modalities: an audio file, a text transcript, and a sequence of sampled image frames, and each utterance is labeled with an emotion label from the set \{\textit{angry}, \textit{happy}, \textit{excited}, \textit{sad}, \textit{frustrated}, and \textit{neutral}\}. The CMU-MOSEI dataset consists of 20,477 annotated utterances. Each utterance consists of three modal inputs including an audio clip, a text transcript, and a sequence of sampled image frames, and each utterance is annotated with multiple emotion labels from the set \{\textit{happy}, \textit{sad}, \textit{angry}, \textit{fearful}, \textit{disgusted}, and \textit{surprised}\}. The statistics of the two adopted datasets are described in Appendix \ref{dataset}. Following prior works~\cite{dai-etal-2021-multimodal}, we use the Accuracy (Acc.) and F1-score to evaluate the models on IEMOCAP. For the CMU-MOSEI dataset, we take the weighted Accuracy (WAcc.) and F1-score as the metrics.

\subsection{Training Details}
We use Adam~\cite{kingma2014adam} as the optimizer. We use the binary cross-entropy loss to train our model on both datasets. The learning rate is set to 1e-4. The epoch number and batch size are set to 40 and 8 respectively. The token number K is set to 256. The max lengths of the visual and textual tokens are set to 576 and 300 respectively. The max lengths of the acoustic tokens are set to 512 and 1024 for IEMOCAP and MOSEI respectively, considering the average duration of the utterances in CMU-MOSEI is much longer than IEMOCAP. Our experiments are run on a Tesla V100S GPU.

For the hyper-parameters of the transformer models, $d$, $k$, $d_h$, and $L$ are 768, 12, 64, and 12 respectively. In practice, we use the pre-trained weights of DeiT to initialize our visual transformer and acoustic transformer. For the visual modality, we adopt MTCNN~\cite{zhang2016joint} to detect faces from the input images. The image resolution of the obtained face image is 128 by 128. We split the face images into 16 $\times$ 16 patches following DeiT~\cite{touvron2021training}. For the acoustic modality, we convert the input audio waveform into a sequence of 128-dimensional log-Mel filterbank (fbank) features computed with a 25ms Hamming window every 10ms and split it into 128 $\times$ 2 patches following AST~\cite{gong21b_interspeech}. We also analyse the effects of different patch construction methods on our proposed model in Appendix \ref{construct}.

\subsection{Baselines}
We compare our proposed model with the two lines of baselines. One line of models adopts a two-phase pipeline. \textbf{LF-LSTM} first uses LSTMs to encode the input features and then fuses them. \textbf{LF-TRANS} uses the transformer models to encode the input features and fuses them for prediction. \textbf{EmoEmbs}~\cite{dai-etal-2020-modality} leverages the cross-modal emotion embeddings for multi-modal emotion recognition. \textbf{MulT}~\cite{tsai-etal-2019-multimodal} utilizes the cross-modal attention to fuse multi-modal features. This line of models takes the extracted features as the inputs. The other baselines are trained in an end-to-end manner. \textbf{FE2E}~\cite{dai-etal-2021-multimodal} first uses two VGG models to encode the visual and acoustic inputs and then utilizes ALBERT~\cite{Lan2020ALBERT} to encode the textual input. \textbf{MESM}~\cite{dai-etal-2021-multimodal} utilizes the cross-modal sparse CNN block to capture the bi-modal interactions based on FE2E. \textbf{FE2E-BERT} and \textbf{MESM-BERT} take BERT as the textual encoder. These four baselines take the raw data as the input.

\subsection{Experimental Results}

We compare our model with baselines, and the experimental results are shown in Table \ref{tab:main_iemocap} and Table \ref{tab:main_mosei}. We can see from the results that our model surpasses all baseline models on two datasets. We attribute the success to strong model capacity and effective feature fusion. Specifically, our model uses three transformer models to model the raw input data, which can capture the global intra-modal dependencies. Moreover, we propose the progressive tri-modal attention and tri-modal feature fusion layer to model the tri-modal feature interactions, which enables the model to predict labels by leveraging the tri-modal information effectively. 

Comparing with the two-phase pipeline models, we find that the end-to-end models can achieve better performance, and we attribute it to that the end-to-end models can extract more task-discriminative features supervised by the target task loss. The comparison between MESM and FE2E shows that although MESM utilizes the bi-modal attention to capture the interactions between textual and acoustic/visual based on FE2E, MESM obtains worse results than FE2E on two datasets. This observation indicates that MESM fails to balance the computation cost with the model performance. Besides, we replace the ALBERT models of the end-to-end baselines with the BERT models for a fair comparison. But FE2E-BERT and MESM-BERT are not clearly superior to FE2E and MSEM, respectively.

\begin{table}[t]
    \caption{Results of the ablation study on the IEMOCAP and MOSEI datasets. The best results are in \textbf{bold}.}
	\centering
	\resizebox{\linewidth}{!}{
	\begin{tabular}{l|ll|ll}
		\hline
		\multicolumn{1}{l|}{\multirow{2}{*}{Models}}  & \multicolumn{2}{c}{IEMOCAP} & \multicolumn{2}{c}{MOSEI}  \\
		  & \multicolumn{1}{c}{Avg.Acc} & \multicolumn{1}{c}{Avg.F1} & \multicolumn{1}{c}{Avg.Acc} & \multicolumn{1}{c}{Avg.F1}  \\
		\hline
		\hline
		\multicolumn{1}{l|}{\multirow{1}{*}{ME2ET}}  & \multicolumn{1}{c}{\textbf{86.5}} & \multicolumn{1}{c}{\textbf{61.3}}  & \multicolumn{1}{c}{\textbf{68.3}} & \multicolumn{1}{c}{\textbf{48.0}} \\
		\multicolumn{1}{l|}{\multirow{1}{*}{w/o Two-pass}}  & \multicolumn{1}{c}{86.2} & \multicolumn{1}{c}{60.7}   & \multicolumn{1}{c}{68.1} & \multicolumn{1}{c}{47.5}\\
        \multicolumn{1}{l|}{\multirow{1}{*}{w/o Feature Fusion}}  & \multicolumn{1}{c}{85.9} & \multicolumn{1}{c}{60.1}  & \multicolumn{1}{c}{66.9} & \multicolumn{1}{c}{47.3} \\
         \multicolumn{1}{l|}{\multirow{1}{*}{w/o Attention}}  & \multicolumn{1}{c}{86.2} & \multicolumn{1}{c}{58.1} & \multicolumn{1}{c}{66.8} & \multicolumn{1}{c}{47.1} \\
        \hline
        \end{tabular}
    }

	\label{tab:main_ablation}
\end{table}

\section{Analysis}

\subsection{Ablation Study}
We conduct the ablation study to distinguish the contribution of each part. There are several variants of our model. \textbf{ME2ET} is our proposed full model. \textbf{ME2ET w/o Attention} does not use the progressive tri-modal attention to capture tri-modal interactions, which uses three transformer models separately to encode the inputs and applies the feature fusion and decision fusion layers to predict the result. \textbf{ME2ET w/o Two-pass} does not perform the attention over the visual tokens in the second pass to obtain the refined visual representation. \textbf{ME2ET w/o Feature Fusion} does not use the tri-modal feature fusion layer.  The result of the ablation study is shown in Table \ref{tab:main_ablation}. We observe that ablating the progressive tri-modal attention hurts the model performance, which demonstrates the importance of our proposed attention. Comparing ME2ET w/o Two-pass with ME2ET, we find that performing the attention over the visual tokens again by leveraging the tri-modal information is useful for model prediction. We also observe that utilizing the proposed tri-modal feature fusion layer can boost the model performance, which can enable ME2ET to model the tri-modal feature interactions at high-level.

\begin{figure}[t]
\centering
\includegraphics[width=\columnwidth]{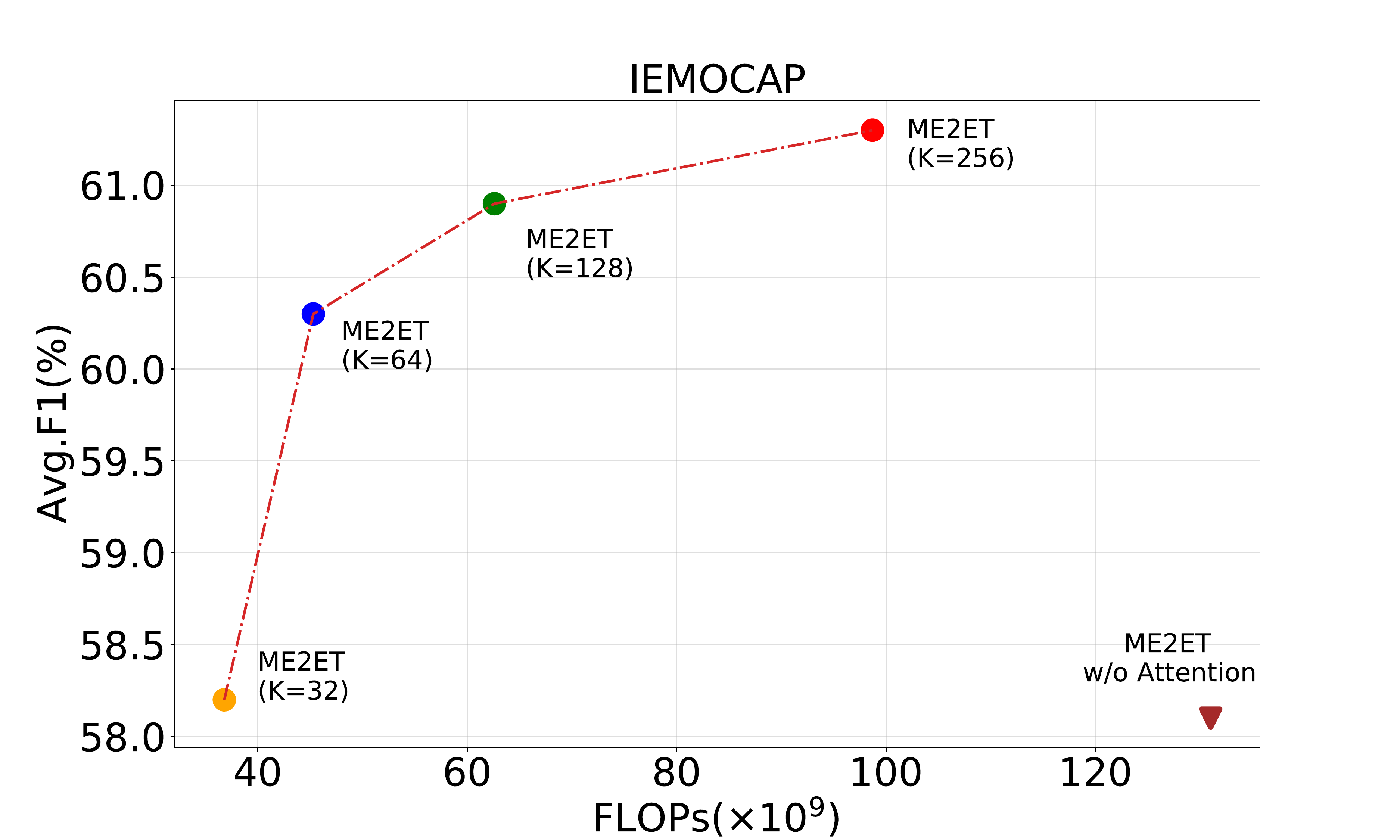} 
\caption{Comparison of ME2ET w/o Attention and ME2ET with different token number $K$.}
\label{flops}
\end{figure}

\begin{figure}[t]
\centering
\includegraphics[width=\columnwidth]{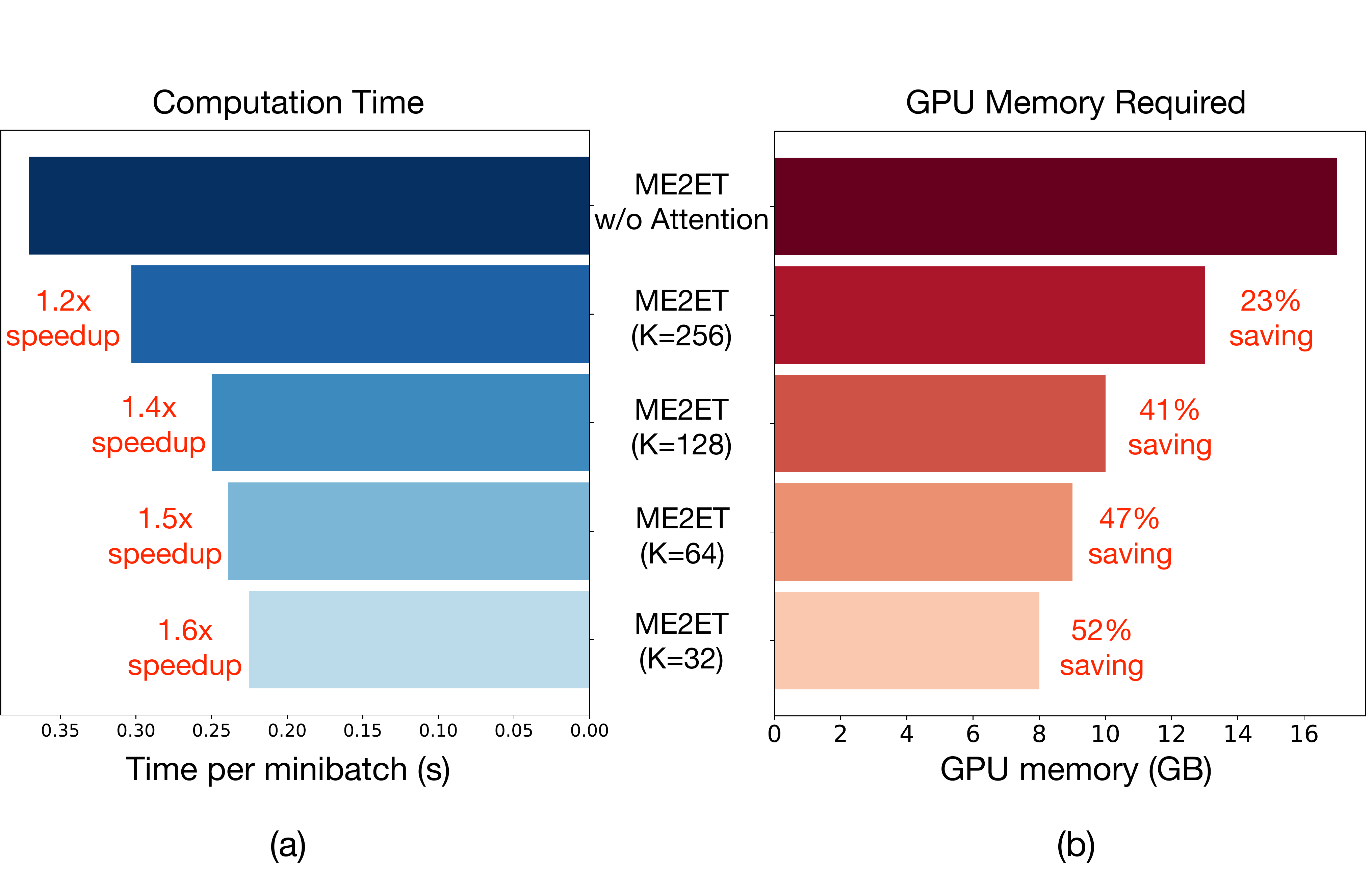} 
\caption{Computation time \textbf{(a)} and GPU memory required \textbf{(b)} for a combined forward and backward pass on IEMOCAP (details in Appendix \ref{evaluate}).}
\label{memory}
\end{figure}

\subsection{Computation and Memory Analysis}
\paragraph{Computation Efficiency.}  Firstly, we adopt the number of float-point operations (FLOPs) as our metric to measure the computation complexity. The performance of ME2ET w/o Attention and ME2ET with different token number $K$ on IEMOCAP is shown in Figure \ref{flops}. Comparing ME2ET with ME2ET w/o Attention, we can find that our models not only significantly reduce the computation complexity but also obtain better results than ME2ET w/o Attention. Specifically, when the token number $K$ is set to 32, our proposed model achieves better performance than ME2ET w/o Attention while only requiring about 3 times 
less computation. We also observe that our model obtains better results as we increase the token number. This is in line with our expectations because a larger $K$ enables the model to address the important information from more perspectives and also increases the model capacity. Secondly, in order to more accurately evaluate the speeds of models, we analyze the computation time performance of ME2ET and ME2ET w/o Attention and show the result in Figure \ref{memory}(a). We can see that ME2ET can obtain 1.2x$\sim$1.6x speedup over ME2ET w/o Attention depending on the selected token number $K$, which indicates that ME2ET is computation-efficient. 

\paragraph{Memory Efficiency.}  We list the GPU memory required for ME2ET and ME2ET w/o Attention in Figure \ref{memory}(b). We can observe that with our proposed tri-modal attention, ME2ET only requires 48\%$\sim$77\%  GPU memory while achieves better performance. Specifically, when the token number $K$ is set to 32, ME2ET only uses 48\% of the memory required by ME2ET w/o Attention. This analysis shows that ME2ET is memory-efficient. 

\begin{figure}[t]
\centering
\includegraphics[width=0.8\columnwidth]{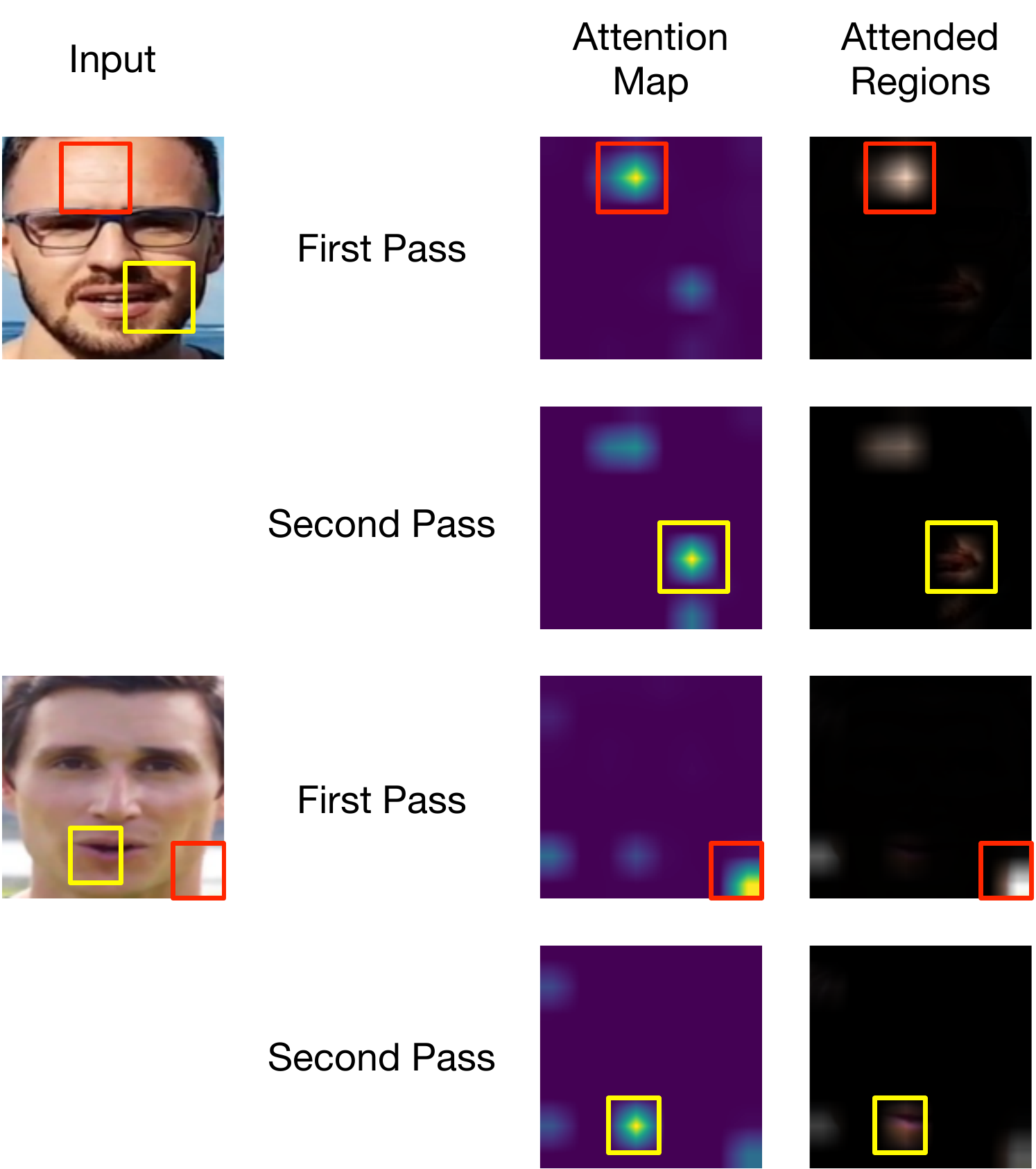} 
\caption{Visualization of the progressive tri-modal attention of the visual modality. From left to right, we show the input image, the attention map, and the attended image regions. We use the red and yellow boxes to highlight the image regions with the larger attention weights in the first pass and second pass respectively.}
\label{case}
\end{figure}

\begin{figure}[t]
\centering
\includegraphics[width=0.95\columnwidth]{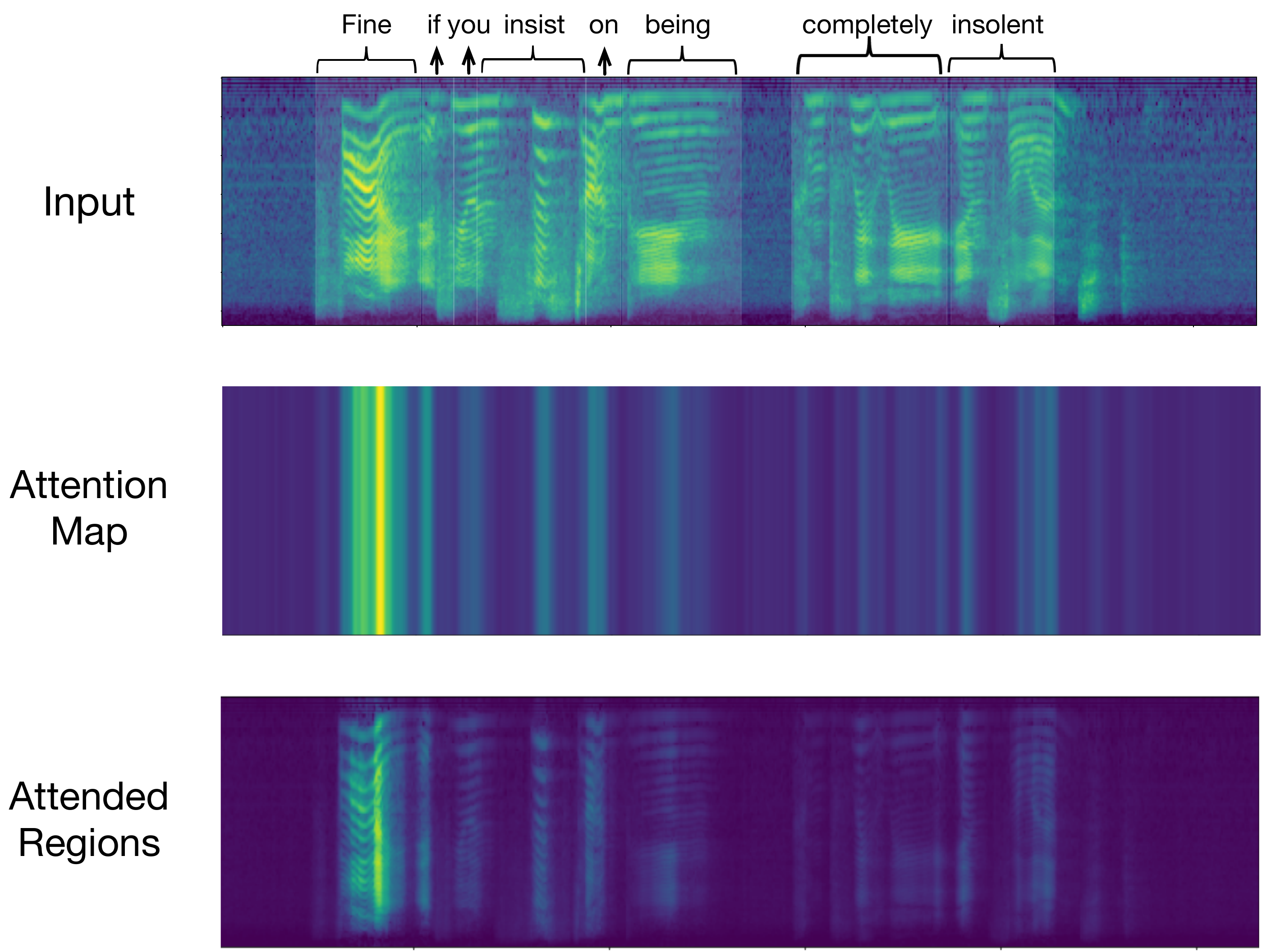} 
\caption{Visualization of the progressive tri-modal attention of the acoustic modality. From top to bottom, we show the input acoustic feature map, the attention map, and the attended regions.}
\label{audio_case}
\end{figure}
\vspace{-0.3cm}

\subsection{Visualization Analysis}

To have an intuitive understanding of our proposed model, we visualize the attention weights of our progressive tri-modal attention. Specifically, we average the attention weights on the input tokens produced by different attention heads and use the obtained attention map to highlight the important regions. Based on the visualization result for the visual modality in Figure \ref{case}, we observe that the proposed progressive tri-modal attention successfully addresses the important image regions. In the first pass, our model only uses the textual and visual information and fails to address the important patches. In the second pass, our model pays more attention to the raised lip corners by capturing the tri-modal feature interactions, which are useful for predicting emotion. As for the acoustic modality, we show an example in Figure \ref{audio_case}. The speaker stresses the word ``Fine" to express her angry emotion and the model places more attention on the region corresponding to the word ``Fine", which is in line with our expectation.

\section{Conclusion}
In this paper, we propose the multi-modal end-to-end transformer (ME2ET), which utilizes the progressive tri-modal attention and tri-modal feature fusion layer to capture the tri-modal feature interactions at the low-level and high-level. We conduct extensive experiments on two datasets to evaluate our proposed model and the results demonstrate its effectiveness. The computation and memory analysis shows that ME2ET is computation and memory-efficient. With the proposed progressive tri-modal attention, ME2ET can achieve better performance by fully leveraging the tri-modal feature interactions, while obtaining 1.2x$\sim$1.6x speedup and saving 23\%$\sim$52\% GPU memory during training. The visualization analysis shows the interpretability of our model, which can successfully address the informative tokens. For future work, we would like to explore building a unified transformer model for multi-modal emotion recognition.

\newpage

\bibliography{anthology,custom}
\bibliographystyle{acl_natbib}

\newpage

\appendix

\section{Complexity Analysis} \label{complexity}
The memory and time complexity of the self-attention and FFN blocks in Transformer are $\mathcal{O}(N^2)$ and $\mathcal{O}(N)$ respectively ~\cite{Kitaev2020Reformer}. The self-attention scales quadratically with the length $N$, which makes it hard to directly apply Transformer to the raw audio and video. Given an utterance length of $x$ seconds, the lengths of the visual and acoustic tokens are $128 \cdot x$ and $100 \cdot x$. Let us take an example. Given an utterance length of 10s, the audio can be transformed into a very long sequence of acoustic tokens of which length is about 1000. If we directly feed them into the acoustic transformer, the computation and memory cost is very high. To address this problem, we propose the progressive tri-modal attention. The main idea of it is to fully leverage the tri-modal information to reduce the length of the input tokens. We utilize it to transform a long sequence ( $N$ tokens) into a fixed-length sequence ($K$ tokens, $K \ll N$) and reduce the memory and time complexity of the self-attention and FFN blocks from $\mathcal{O}(N^2)$ and $\mathcal{O}(N)$ to $\mathcal{O}(K^2)$ and $\mathcal{O}(K)$ respectively.

\section{Datasets} \label{dataset}

We show the statistics of the two adopted datasets in Table \ref{tab:dataset}. The average duration of the utterances in CMU-MOSEI is much longer than IEMOCAP. Thus, we set the max lengths of the input acoustic tokens to 1024 for MOSEI and 512 for IEMOCAP. As for the visual modality, we only sample some key image frames from the video following the previous work~\cite{dai-etal-2021-multimodal} and set the max length of the input visual tokens to 576 for both datasets.

\begin{table}[h]
	\centering
	\caption{Statistics of the IEMOCAP and CMU-MOSEI datasets.}
	\resizebox{\linewidth}{!}{
	\begin{tabular}{l|l|l|lll}
		\hline
		\multicolumn{1}{l|}{\multirow{1}{*}{Dataset}}  & \multicolumn{1}{c|}{Label}  & \multicolumn{1}{c|}{Avg. duration(s)}  &\multicolumn{1}{c}{Train} & \multicolumn{1}{c}{Valid} & \multicolumn{1}{c}{Test}  \\
		\hline
		\multicolumn{1}{l|}{\multirow{6}{*}{IEMOCAP}}  & \multicolumn{1}{c|}{Anger} & \multicolumn{1}{c|}{4.51} & \multicolumn{1}{c}{757} & \multicolumn{1}{c}{112} & \multicolumn{1}{c}{234}  \\

		 & \multicolumn{1}{c|}{Excited}  & \multicolumn{1}{c|}{4.78}  & \multicolumn{1}{c}{736} & \multicolumn{1}{c}{92} & \multicolumn{1}{c}{213}  \\
		 & \multicolumn{1}{c|}{Frustrated}  & \multicolumn{1}{c|}{4.71} & \multicolumn{1}{c}{1,298} & \multicolumn{1}{c}{180} & \multicolumn{1}{c}{371}  \\
		
		  & \multicolumn{1}{c|}{Happy}  & \multicolumn{1}{c|}{4.34} & \multicolumn{1}{c}{398} & \multicolumn{1}{c}{62} & \multicolumn{1}{c}{135}  \\
		  & \multicolumn{1}{c|}{Neutral}  & \multicolumn{1}{c|}{3.90} & \multicolumn{1}{c}{1,214} & \multicolumn{1}{c}{173} & \multicolumn{1}{c}{321}  \\
		  & \multicolumn{1}{c|}{Sad}  & \multicolumn{1}{c|}{5.50}  & \multicolumn{1}{c}{759} & \multicolumn{1}{c}{118} & \multicolumn{1}{c}{207}  \\
		  \hline
		  
		  \multicolumn{1}{l|}{\multirow{6}{*}{CMU-MOSEI}}  & \multicolumn{1}{c|}{Anger}  & \multicolumn{1}{c|}{23.24}  & \multicolumn{1}{c}{3,267} & \multicolumn{1}{c}{318} & \multicolumn{1}{c}{1,015}  \\

		 & \multicolumn{1}{c|}{Disgusted}  & \multicolumn{1}{c|}{23.54 } & \multicolumn{1}{c}{2,738} & \multicolumn{1}{c}{273} & \multicolumn{1}{c}{744}  \\
		 & \multicolumn{1}{c|}{Fear}  & \multicolumn{1}{c|}{28.82}  & \multicolumn{1}{c}{1,263} & \multicolumn{1}{c}{169} & \multicolumn{1}{c}{371}  \\
		
		  & \multicolumn{1}{c|}{Happy}  & \multicolumn{1}{c|}{24.12}  & \multicolumn{1}{c}{7,587} & \multicolumn{1}{c}{945} & \multicolumn{1}{c}{2,220}  \\
		  & \multicolumn{1}{c|}{Sad}  & \multicolumn{1}{c|}{24.07} & \multicolumn{1}{c}{4,026} & \multicolumn{1}{c}{509} & \multicolumn{1}{c}{1,066}  \\
		  & \multicolumn{1}{c|}{Surprised}  & \multicolumn{1}{c|}{25.95} & \multicolumn{1}{c}{1,465} & \multicolumn{1}{c}{197} & \multicolumn{1}{c}{393}  \\
		
        \hline
        \end{tabular}
    }
	\label{tab:dataset}
\end{table}

\section{Effect of Patch Construction} \label{construct}

To analyze the impact of patch construction on our model, we perform experiments with different patch construction methods and report the results in Table \ref{tab:face_size}. We keep the patch size at 16 $\times$ 16 for the visual transformer and evaluate our model with different face image sizes from 48 $\times$ 48 to 128 $\times$ 128. As shown in Table \ref{tab:face_size}, there is a slight drop in the evaluation performance when we change the image size from 48 $\times$ 48 to 64 $\times$ 64. A possible reason is that the number of the input face images decreases as we increase the face image size since we fix the max patch number for batch computation. On the other hand, increasing the image size enables our model to capture more detailed information. 
For this reason, when we set the image size to 128 $\times$ 128, the model achieves better results. For the acoustic transformer, we evaluate two patch construction methods. One is splitting the input spectrogram into 16 $\times$ 16 square patches. But we consider constructing patches in this way may lose the timing information. Hence, we split the spectrogram into 128 $\times$ 2 rectangle patches. The results demonstrate the effectiveness of our method.

\begin{table}[h]
    \caption{Experimental results with different patch construction methods on the IEMOCAP dataset. The best results are in \textbf{bold}.}
	\centering
	\resizebox{\linewidth}{!}{
	\begin{tabular}{ll|ll}
		\hline
		\multicolumn{1}{c}{\multirow{1}{*}{Face Image Size}}  & \multicolumn{1}{l|}{\multirow{1}{*}{Time Split}}  & \multicolumn{1}{c}{Avg.Acc} & \multicolumn{1}{c}{Avg.F1}  \\
		\hline
		\hline
		\multicolumn{1}{c}{\multirow{1}{*}{48 $\times$ 48}}  & \multicolumn{1}{c|}{\multirow{1}{*}{True}}  & \multicolumn{1}{c}{86.2} & \multicolumn{1}{c}{60.4}  \\
		\multicolumn{1}{c}{\multirow{1}{*}{64 $\times$ 64}}  &  \multicolumn{1}{c|}{\multirow{1}{*}{True}}  &\multicolumn{1}{c}{85.9} & \multicolumn{1}{c}{59.4}  \\
        \multicolumn{1}{c}{\multirow{1}{*}{128 $\times$ 128}}  & \multicolumn{1}{c|}{\multirow{1}{*}{False}}  & \multicolumn{1}{c}{85.9} & \multicolumn{1}{c}{58.9}  \\
		\multicolumn{1}{c}{\multirow{1}{*}{128 $\times$ 128}}  & \multicolumn{1}{c|}{\multirow{1}{*}{True}}  & \multicolumn{1}{c}{\textbf{86.5}} & \multicolumn{1}{c}{\textbf{61.3}}  \\
        \hline
        \end{tabular}
    }
	\label{tab:face_size}
\end{table}

\section{Computation and GPU Memory Evaluation}\label{evaluate}
We benchmark by running a forward and backward pass. Each measurement is an average over 100 runs on an Tesla V100S 32GB. The code is implemented in the PyTorch framework v1.8.1\cite{paszke2019pytorch}. The batch size is set to 8. The max lengths of the acoustic, visual, and textual tokens are set to 512, 576, and 300 respectively.

\end{document}